# MaskFlow: Object-Aware Motion Estimation

Aria Ahmadi
aria.ahmadi@imgtec.com
David R. Walton
david.walton@imgtec.com
Tim Atherton
tim.atherton@imgtec.com
Cagatay Dikici
cagatay.dikici@imgtec.com

Imagination Technologies
Kings Langley
Hertfordshire, UK

Abstract

We introduce a novel motion estimation method, MaskFlow, that is capable of estimating accurate motion fields, even in very challenging cases with small objects, large displacements and drastic appearance changes. In addition to lower-level features, that are used in other Deep Neural Network (DNN)-based motion estimation methods, MaskFlow draws from object-level features and segmentations. These features and segmentations are used to approximate the objects' translation motion field. We propose a novel and effective way of incorporating the incomplete translation motion field into a subsequent motion estimation network for refinement and completion. We also produced a new challenging synthetic dataset with motion field ground truth, and also provide extra ground truth for the object-instance matchings and corresponding segmentation masks. We demonstrate that MaskFlow outperforms state of the art methods when evaluated on our new challenging dataset, whilst still producing comparable results on the popular FlyingThings3D benchmark dataset.

## 1 Introduction

The goal of motion estimation is to determine how pixels move from a reference frame to a target frame. Several methods have been proposed to address this problem some of which use Deep Neural Networks (DNN-based methods) and the rest which we refer to as classical methods. Most of classical methods work based on penalizing the deviation from the feature constancy assumption. DNN-based methods, during supervised training, learn to estimate a motion field that is the closest to a ground truth and exploit the learned model during the inference. A current major challenge is that objects change in appearance due to rotation, lighting changes etc. Although methods have become increasingly robust as the field has advanced, state of the art methods still struggle in very challenging scenarios with small objects undergoing large displacements and drastic appearance changes.

Early classical methods exploited intensity values as features [9, 18]. Later classical methods increased robustness by using higher-level representations for matching correspondences [3, 4, 25]. Current DNN-based methods learn to generate representations that are even more robust to these variations [1, 6, 12, 14, 15, 23, 27, 30]. Recent DNN-based methods have begun to outperform classical methods both in terms of computational efficiency and accuracy. State of the art methods [14, 27] use a multi-scale approach which results in some shortcomings, particularly where small objects are displaced by a large distance between the two frames. In these cases, objects might fall outside the limited spatial range at the higher spatial resolutions, whereas at lower spatial resolutions information about the small objects may be lost after multiple down-samplings. In addition, these features are still sometimes not robust enough to represent objects which change dramatically in appearance due to large rotations, occlusion etc. Nevertheless, humans are often still able to understand the motion occurring in such examples due to their higher-level understanding of the objects in the scene. We hypothesise that by identifying instances of objects and matching them between the two frames, they can resolve these challenging cases. With this intuition, we developed MaskFlow, a motion estimation method which combines object-level representations and matching with a state of the art multi-scale DNN-based method [27] to handle these difficult cases, but retain the accuracy of existing methods when estimating smaller motions.

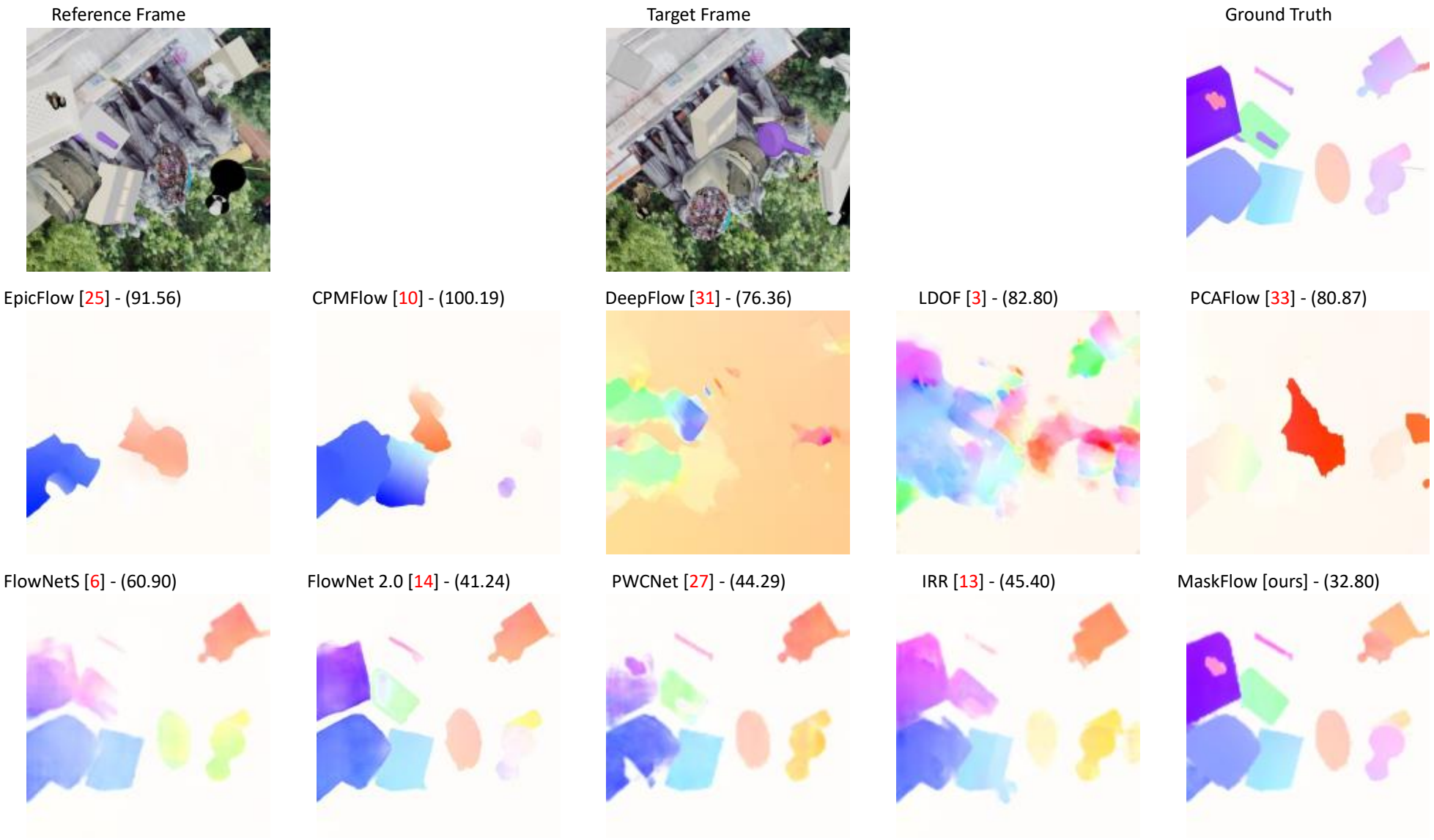

Figure 1: The input, ground truth, and output of state of the art classic and DNN-based methods on one sample from the test split of our MaskFlow dataset - the parentheses contain the Average Endpoint Error (AEE) in pixels. For a fair comparison, we have trained all DNN-based methods under the same training schemes, $S_{long}$ and $S_{fine}$, proposed in [14].

We propose MaskFlow that is, to the best of our knowledge, the first DNN-based method that exploits object-level features for motion estimation. MaskFlow estimates motion in two steps: 1) estimation of translation-only motion field for objects 2) refinement and completion to form the full motion field. In step one we use a modified Mask R-CNN [8] to identify and segment objects in the two frames. We modify the Mask R-CNN class estimation branch to output class-agnostic features and adopt a loss function to make these features as discriminative as possible for later object matching. We also add a novel mask score branch which further assists matching. We then use a novel technique to filter and match the candidates proposed by Mask R-CNN, allowing us to generate a translation motion field. In step two, we propose a new way of completing and refining the translation motion field using a modified PWC-Net [27], producing a final estimate of the motion field. We show that our two-step approach performs well in estimating the motion of objects undergoing drastic appearance changes. Our experimental results show that we outperform all other classic and DNN-based methods when tackling our challenging dataset containing such appearance changes. However, we still perform comparably on an easier benchmark dataset [20].

We plan to release our code, trained models and rendered dataset to facilitate future research into challenging motion estimation problems.

## 2 Related Work

Our proposed method receives as input two consecutive frames and estimates the motion field between them. Here we discuss such two-frame motion estimation methods, but many multiframe methods also exist [17, 19, 22, 24, 34]. Early classical motion estimation methods such as [9, 18] were capable of producing motion fields in a specific setting, but struggled to correctly estimate large translations, and were not very robust to more severe changes in object appearance. Some more recent methods such as DeepFlow [31], EpicFlow [25] and CPMFlow [10] try to address these issues. They first estimate sparse matches using a more robust method such as Deepmatching [32] and then densify it to a full motion field.More recent methods have made use of trained neural networks, some of these are outlined below.

DNN-based methods have been trained both in supervised and unsupervised ways. Here we focus on methods trained in a supervised way similar to MaskFlow, but many successful unsupervised methods exist [1, 2, 15, 21, 30]. FlowNet [6] introduced the first successful end-to-end DNN-based method for motion estimation, pioneering the use of correlation layers and large synthetic datasets for training CNN-based motion estimation networks. FlowNet was substantially faster than existing methods such as EpicFlow, but still lower in accuracy. FlowNet 2.0 [14]) used a revised architecture combining multiple modified instances of FlowNet and was able to outperform EpicFlow in accuracy and performance. SPyNet [23] made use of an image pyramid method used by many classical motion estimation methods, applying a DNN at each pyramid level to estimate the flow field. PWC-Net [27, 28] combined such an image pyramid with correlation and refinement layers similar to FlowNet. More recent methods have built upon PWC-Net, adding occlusion detection and temporal connections [22] and reducing the model size by sharing decoders [13].

# 3 Method

The input for motion estimation is a pair of frames; the reference and the target, denoted by $I_1(x,y)$ and $I_2(x,y)$ respectively. The goal of motion estimation is to find $D\hat{} : [0,h] \times [0,w] \rightarrow \mathbb{R}^2$, an approximation to the dense ground truth motion field $D(x,y)$ that describes how pixels are displaced from $I_1$ to $I_2$. Here $h$ and $w$ are the height and width of the input frames respectively. Each entry of $D(x,y)$ gives an offset between the surface point at $(x,y)$ in $I_1$ and its correspondence in $I_2$, in image space.

We assume that the motion of objects can be adequately modelled as a coarse, large-scale translation followed by fine-scale translation and rotation. We attempt to simplify estimation of a full motion field by first estimating objects' rigid translation motion field, $D_t$, then let a refinement network complete it. At each object pixel $D_t$ is equal to the displacement of the centroids of the pixels containing the object between the reference and target frames.

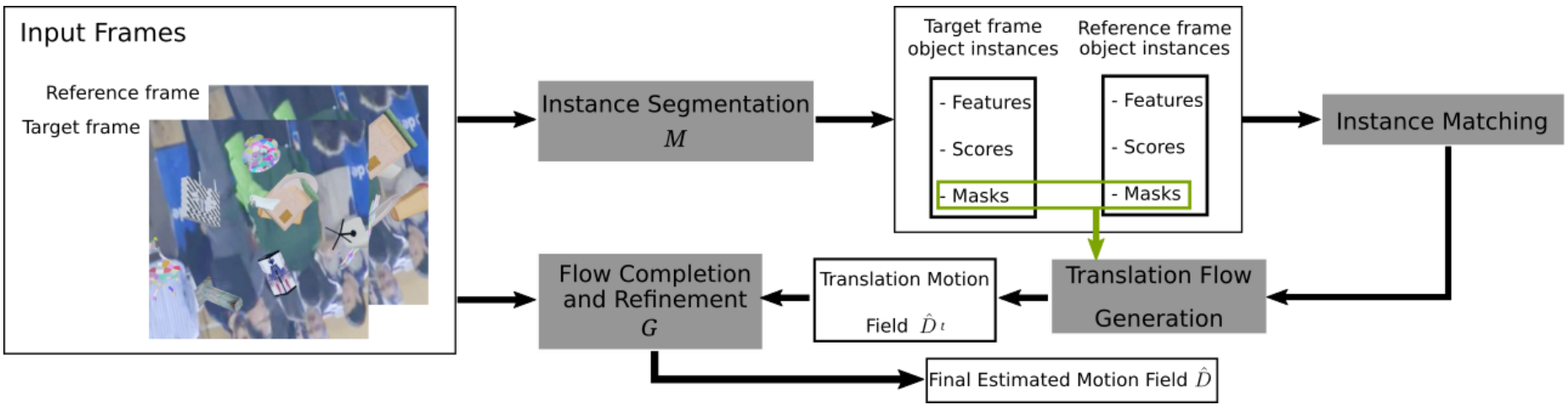

Figure 2: Overview of our proposed motion estimation method, MaskFlow.

## 3.1 Method Overview

An overview of our proposed method, MaskFlow, is shown in figure 2. Briefly, we first segment object instances in the two frames in an instance segmentation stage. These instances are matched between the two frames in an instance matching stage. The matched instances and their masks are then used to generate an approximation of the translation motion field, $\widehat{D}_t$. Finally, $\widehat{D}_t$ is used as input to a completion and refinement stage, which generates the final motion field, $\widehat{D}$.

We estimate the translation motion field $\widehat{D}_t$ by segmenting and matching object instances. To do this, we train a modified version of Mask R-CNN [8]. This identifies object instances $c \in 1,\ldots,N$ in $I_1$ and $I_2$ and calculates a bounding box, $bb_c$, a binary mask, $M_c$, an objectness score, $O_c$, a (novel) mask score, $S_c$, and a representative feature vector, $F_c$ for each. Objects are matched between the two frames based on the feature vectors, masks and scores. The offsets between the centroids of the masks in the reference frame and the centroids of their matched corresponding masks in the target frame are then used to generate $\widehat{D}_t$.

$\widehat{D}_t$ is then completed using a refinement network $G$. This gives the final estimation of the motion field $\widehat{D} = G(I_1, I_2, \widehat{D}_t)$. This process is illustrated in figure 2.

## 3.2 Instance Segmentation

Our modified Mask R-CNN network, $M$, identifies a number of object instances and corresponding masks for frames 1 and 2. The architecture of $M$ is shown in figure 3. In the original Mask R-CNN a Region Proposal Network (RPN) is first used to generate a series of object proposals, each consisting of a bounding box. A small feature map is extracted from each of these boxes and passed through a second "head" network, branches of which produce masks and object classes. We use the ResNet50 FPN architecture from [8]. In $M$ the branch which in Mask R-CNN would be used to extract object categories is modified to extract a $256D$ feature vector; more detail is given in section 3.2.1 below. We further add an extra branch which extracts a novel mask quality score, described in section 3.2.2.

In order to use these to generate the translation motion field $\widehat{D}_t$, we first remove lower quality masks and then identify which pairs of instances in frames 1 and 2 correspond to the same real object. We do so using the matching process detailed in section 3.3 below.

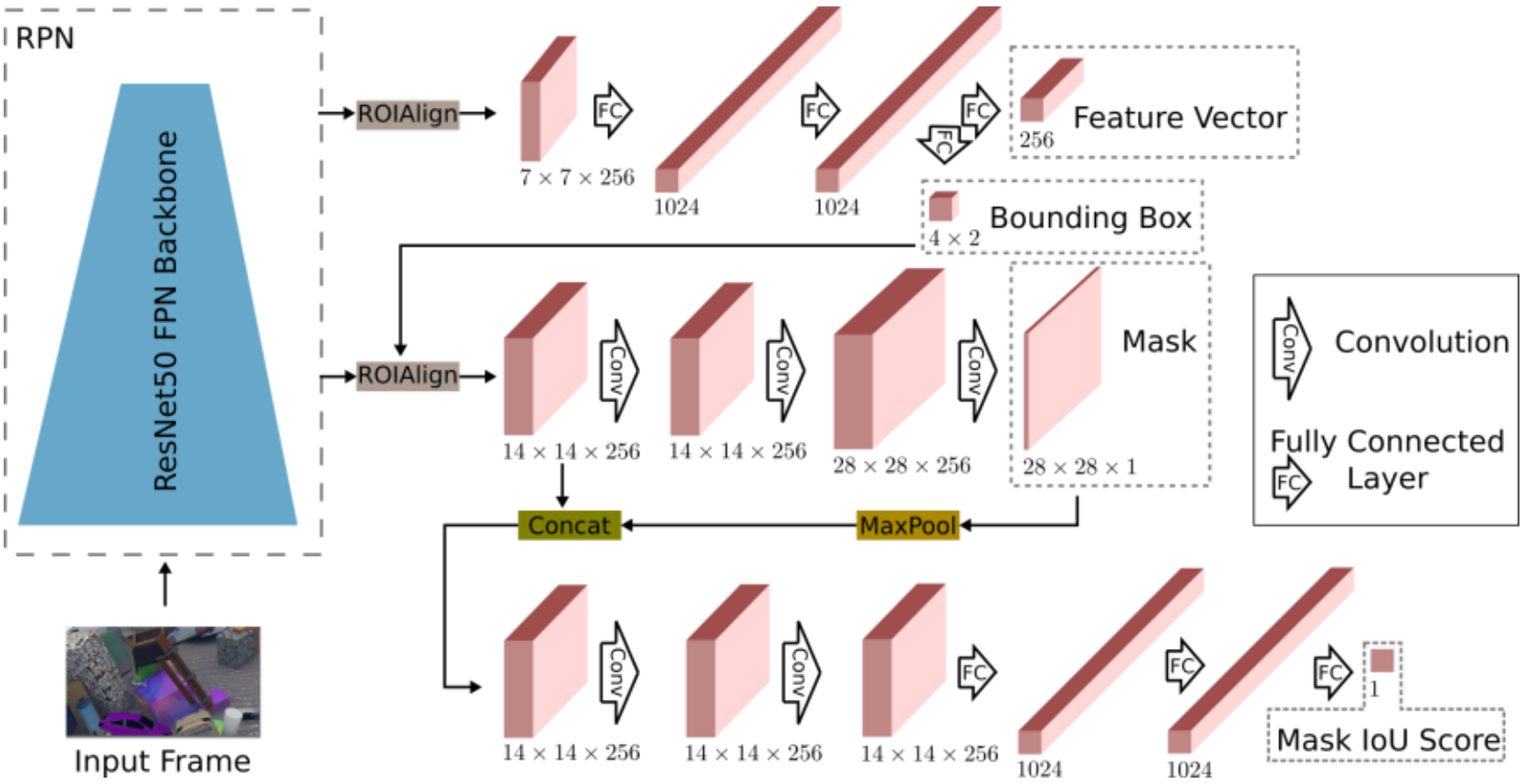

Figure 3: Architecture of our modified Mask R-CNN instance segmentation network.

### 3.2.1 Class-agnostic Object Features

Let $\Omega$ be the set of ground truth objects in the input frame pair. The original Mask R-CNN method finds, for each object $\omega \in \Omega$, $c_1$ and $c_2$ candidates in frames 1 and 2, respectively. Rather than outputting a class prediction vector [8], our modified instance segmentation network, for each predicted instance $c$, outputs a feature vector $fc \in \mathbb{R}^{256}$.

The loss we use to train the feature generating branch is an adapted Triplet Loss function [26]. For each $\omega \in \Omega$, from the candidates identified by the network, we pick a random feature vector $f_\omega$ to represent it. We then select random subsets of matching features $F_m$ and non-matching features $F_n$. All $f \in F_m$ are features corresponding to other candidates of object $\omega$, and all $f \in F_n$ correspond to objects other than $\omega$. We select as many candidates as possible, up to a maximum of 9 (i.e. $|F_m| \leq 9$ and $|F_n| \leq 9$)[1]. The similarity and dissimilarity losses $L_{Fs}$ and $L_{Fd}$ are defined as follows:

$$L_{Fs}(\omega) := \frac{1}{|F_m|} \sum_{f \in F_m} |f_w - f|_1 \qquad L_{Fd}(\omega) := \frac{1}{|F_n|} \sum_{f \in F_n} |f_\omega - f|_1 \tag{1}$$

The overall feature loss function we minimize is:

$$L_F := \frac{1}{|\Omega|} \sum_{\omega \in \Omega} \max\left(L_{Fs}(\omega) - L_{Fd}(\omega) + \epsilon, 0\right) \tag{2}$$

In our implementation, we set the margin $\epsilon := 2$.

### 3.2.2 Mask Confirmation Branch

Our proposed mask confirmation branch learns to estimate the quality of the masks, that are generated by the mask branch of the Mask R-CNN. It receives as input the concatenation of the input and output of the mask branch. During the training it learns to output a mask score $s_c$ for each candidate object instance $c$. The mask confirmation loss term is defined as:

$$L_S := \frac{1}{|C|} \sum_{c \in C} \left| IoU(m_c, m_{gt} - s_c \right|_1 \tag{3}$$

Here, $C$ is the set of all candidate object instances, $m_c$ is the estimated mask of candidate $c$ and $m_{gt}$ is the ground truth mask. The mask score learns to approximate the Intersection over Union ($IoU$) of the masks, which provides a measure of mask accuracy.

The idea of using mask score to remove false positives is similar to the approach used in [11]. However, [11] use the classification score for the predicted class. Here, because the segmentation is class agnostic, we use the $IoU$ of the estimated and ground truth masks.

The architecture of this branch consists of 2 3x3 convolutional layers followed by 3 fully connected layers, as shown in figure 3.

### 3.2.3 Instance Segmentation Training Loss

When training the instance segmentation network, we minimise the sum of these losses $L_I := L_F + L_S + L_B + L_M + L_R$. Here, $L_B$ and $L_M$ are the losses applied to the bounding box and mask outputs, and $L_R$ is the loss used to train the $RPN$. These are unchanged from Mask R-CNN [8].

[1] Limiting $|F_m|$ and $|F_n|$ to 9 constrains the complexity of computing the loss. This proved essential in order to train the network in a reasonable amount of time.

## 3.3 Object Instance Matching

Our modified Mask R-CNN finds a set of candidate object instances $C_r$ in the reference frame and a set $C_t$ in the target frame. Typically, an excess of candidates is found. We need to identify the best candidate corresponding to each object instance in each frame, and find the matches between them. We do so using a five-step process.

In step 1, we apply the clustering algorithm HDBScan [5] to all the candidates in $C_r \cup C_t$. We cluster based upon the feature vectors $f_c$ extracted for each object instance $c$ during the instance segmentation stage. This produces initial clusters $C_i \subseteq C_r \cup C_t$.

Following this, in step 2 we prune candidate instances whose objectness, mask score or mask area fails to meet a threshold. That is, any candidates $c$ where $o_c < t_o$ or $s_c < t_c$ or $a(m_c) < t_a$, where $a$ calculates the area of a mask.

In step 3, we consider instances from the reference and target frames separately; i.e. for each cluster$C_i$ we divide it into clusters $C_{r,i} := C_i \cap C_r$ and $C_{t,i} := C_i \cap C_t$. Within each frame, it is still possible that two similar-looking objects from different parts of the frame form part of the same cluster. To resolve this, we split clusters from each frame if the centroids of the instances are too far apart, using the elbow method [29] for k-means. This provides a good quality set of clusters $\hat{C}_{r,i}, \hat{C}_{t,j}$ in the reference and target frames respectively.

In step 4 we try to identify the single best object instance $c_i$ from each cluster $\hat{C}_{r,i}$, and the best $c_j$ from each cluster $\hat{C}_{t,j}$. We do so by selecting the instance with the largest mask area in each cluster.

At this point, we have sets of good quality object instances $c_i$ and $c_j$ in the reference and target frames, but given the possibility that new clusters may appear in step 3, we need to provide new correspondences between the instances. In step 5, we match these best instances using a simple greedy algorithm. We first find the distance, $d(i,j) := \left| f_{c_i}, f_{c_j} \right|_2$, for all pairs $i, j$. Here $f_c$ is the $256D$ feature vector for candidate $c$. We then assign the matches, proceeding from the lowest $d$ and continuing until all the candidates from at least one of the frames are matched or the distances $d$ exceed a threshold $t_d$. We found our thresholds and clustering parameters via a grid search, minimising the AEE of the translation motion field (computed as outlined below). These parameters are given in the supplementary material.

## 3.4 Translation Motion Field Generation

Once matches have been identified, generating the translation motion field estimate $\widehat{D}_t$ is straightforward. We initialise the motion field to zero. For each matched real object $\omega$, we find a translation motion vector by subtracting the object's location in the reference frame from that in the target frame, i.e. $\widehat{D}_t(\omega) = k(m_{\omega,r}) - k(m_{\omega,t})$.

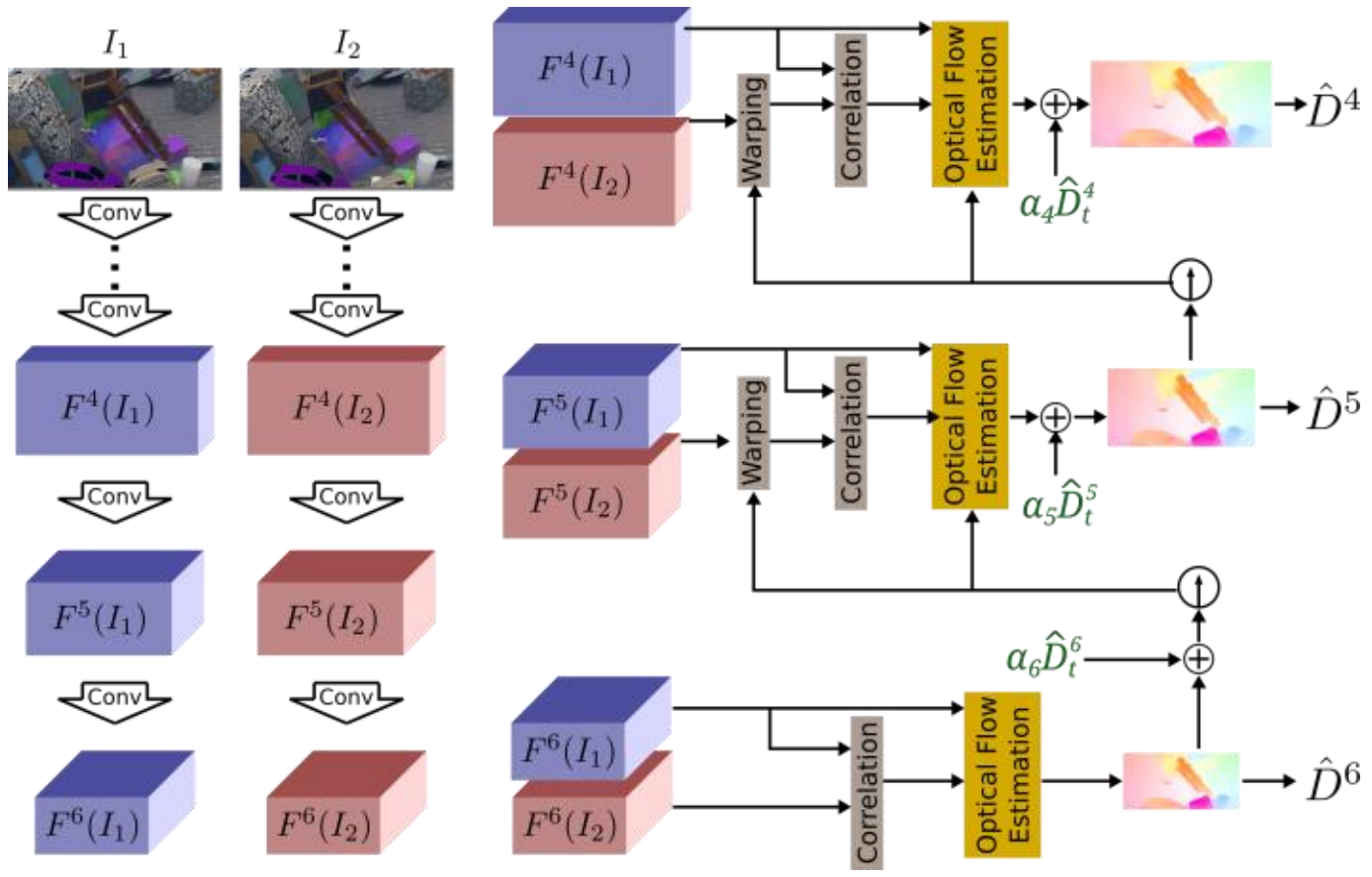

Figure 4: The architecture of our flow refinement process (note that only resolutions 4, 5 and 6 are shown in this illustration for clarity). The notations in green indicate where $\widehat{D}_t$ contributes to the final estimation.

Here $m_{\omega,r}$ and $m_{\omega,t}$ are the masks of $\omega$ in the reference and target frames, and $k(m)$ calculates the centroid of the mask $m$. We then set all flow values in $m_{\omega,r}$ to this flow vector, repeating this for each matched object.

## 3.5 Flow Completion and Refinement

In this implementation, we have chosen the state of the art PWC-Net [27] architecture for the refinement network $G$. PWC-Net is a spatial pyramid network, estimating optical flow at a low resolution first, and then successively upsampling and refining at higher resolutions until a full-resolution motion field is obtained. The network operates at 6 resolutions; in the following, we number these 1 through 6 with 6 being the lowest resolution.

We modify PWC-Net architecture to incorporate $\widehat{D}_t$ at resolutions 6, 5 and 4 of the pyramid, as depicted in figure 4. We empirically found that providing $\widehat{D}_t$ at resolutions 1, 2 and 3 did not improve results. At each resolution $s$, a downsampled $\widehat{D}_t^s$ is added, weighted by a value $\alpha_s$. These $\alpha$ control the contribution of $\widehat{D}_t$ at each resolution independently, and are trained together with the other network weights to optimise the contribution of the translation motion field at each resolution. When training, we use PWC-Net's multi-scale loss.

# 4 Dataset

In order to train our approach, we required a sufficiently large and diverse dataset, containing objects undergoing large translations and varying significantly in appearance due to large rotations or occlusions. For this reason, we rendered our own dataset, which we call the MaskFlow dataset.

We adopt a similar approach to FlyingThings3D [20] but with some modifications as outlined below. We use 16,000 3D objects downloaded from Archive3D[2]: 12,000 used in rendering of training split and 4,000 for test. We animate them with random translations and rotations of 50-200 pixels and 40-80 degrees respectively, in front of a planar background. These translations and rotations are substantially larger than those in FlyingThings3D. The objects and backgrounds are randomly textured with procedural textures and images from a number of datasets [3]. In total, we rendered 28,000 and 3,550 images of size 512x512 pixels for training and test sets, respectively. We produced colour images, object segmentations and ground truth for $D$ and $D_t$. Rendering was performed using a realistic path-tracing renderer
[7].

# 5 Experimental Results

We quantitatively compare our novel approach to the highest-capacity state of the art methods. We compute AEE over images as a whole, and also compute histograms of AEE over ground truth motion magnitude ($d$). We test all methods on the FlyingThings3D dataset and our own rendered dataset, which have ground truth object instance masks. We note that our approach requires object instance masks as a component of the training data, and these are not available for the majority of optical flow datasets.

We train all DNN-based methods in tables 1 and 2 on FlyingChairs using the $S_{long}$ training scheme, and then fine tune on the particular dataset (i.e. FlyingThings3D or the MaskFlow dataset) using the $S_{fine}$ scheme, as described in [14]. We note that $D_t$ is not provided in FlyingThings3D, and so we approximate it using the index maps for the parameter search of our matching technique. To the best of our knowledge, other benchmark datasets do not provide these index maps.

## 5.1 MaskFlow Dataset

The $AEEs$ for comparison with a number of state of the art approaches on our dataset are shown in table 1. MaskFlow outperforms the next best method by over 5.5 and 8.5 pixels in the training and test set respectively. FlowNet 2.0 performs slightly better for small motions; this is likely due to the dedicated small-displacement network unique to FlowNet2.0 which our method lacks. We outperform PWC-Net significantly, showing the benefit of empowering it with our object-aware paradigm. We observe the largest improvement on very large displacements.

[2] http://archive3d.net/

[3] OpenImages [16] and http://www.imageafter.com/

| | Method | Training | | | | | Test | | | | |
|---|---|---|---|---|---|---|---|---|---|---|---|
| | | $AEE$ | $d_{0-10}$ | $d_{10-60}$ | $d_{60-140}$ | $d_{140+}$ | $AEE$ | $d_{0-10}$ | $d_{10-60}$ | $d_{60-140}$ | $d_{140+}$ |
| Classical | EpicFlow [25] | 98.01 | 15.39 | 40.43 | 120.30 | 285.25 | 99.29 | 16.22 | 40.91 | 120.85 | 286.90 |
| | CPMFlow [10] | 98.80 | 14.04 | 38.21 | 127.24 | 292.96 | 100.14 | 15.15 | 38.54 | 127.97 | 295.19 |
| | DeepFlow [31] | 98.62 | 15.73 | 40.14 | 116.82 | 288.99 | 100.52 | 17.18 | 41.25 | 117.66 | 291.21 |
| | LDOF [3] | 88.97 | 9.99 | 33.08 | 109.63 | 270.04 | 89.76 | 10.20 | 33.21 | 109.35 | 271.10 |
| | PCAFlow [33] | 89.20 | 9.10 | 32.97 | 111.90 | 271.46 | 90.04 | 9.22 | 33.06 | 111.99 | 272.77 |
| DNN-based | FlowNetS [6] | 50.60 | 5.61 | 7.75 | 106.78 | 169.71 | 53.61 | 5.76 | 8.02 | 111.79 | 179.71 |
| | FlowNet 2.0 [14] | 36.92 | **3.51** | **3.71** | 80.52 | 128.14 | 40.57 | **3.65** | **3.87** | 87.47 | 140.92 |
| | PWC-Net [27] | 47.09 | 5.30 | 6.17 | 96.72 | 160.79 | 47.53 | 5.38 | 6.28 | 95.89 | 161.65 |
| | IRR [13] | 48.59 | 6.74 | 6.97 | 102.98 | 162.96 | 49.06 | 6.78 | 6.99 | 102.18 | 164.22 |
| | MaskFlow(ours) | **31.03** | 4.91 | 4.99 | **75.37** | **100.46** | **31.87** | 5.10 | 5.06 | **75.28** | **103.23** |

Table 1: Comparison of AEE on our MaskFlow dataset.

## 5.2 FlyingThings3D

The $AEEs$ for comparison with a number of state of the art approaches on the whole FlyingThings3D dataset [20] are shown in table 2. In contrast to the literature, we only excluded flow fields with a maximum flow value exceeding 2000 pixels. We note that our $AEE$ on the test set is within 2 pixels of the best-performing state of the art methods. The $AEE$ on the training and test sets fluctuated significantly during training making it more difficult to precisely determine which methods were best overall. To the best of our knowledge, earlier papers have not reported results on this dataset.

| | Method | Training | | | | | Test | | | | |
|---|---|---|---|---|---|---|---|---|---|---|---|
| | | $AEE$ | $d_{0-10}$ | $d_{10-60}$ | $d_{60-140}$ | $d_{140+}$ | $AEE$ | $d_{0-10}$ | $d_{10-60}$ | $d_{60-140}$ | $d_{140+}$ |
| Classical | EpicFlow [25] | 11.33 | 2.27 | 5.20 | 23.86 | 94.04 | 12.22 | 2.07 | 5.50 | 24.88 | 111.58 |
| | CPMFlow [10] | 12.57 | 2.77 | 7.36 | 22.87 | 95.34 | 11.96 | 2.07 | 5.21 | 25.18 | 108.32 |
| | DeepFlow [31] | 11.41 | 2.52 | 5.28 | 22.30 | 97.96 | 11.22 | 2.49 | 5.01 | 22.02 | 101.79 |
| | LDOF [3] | 18.82 | 2.08 | 4.73 | 43.27 | 199.65 | 18.65 | 2.21 | 4.92 | 43.49 | 200.17 |
| | PCAFlow [33] | 14.47 | 2.35 | 7.03 | 30.60 | 118.17 | 15.30 | 2.23 | 7.42 | 33.03 | 129.40 |
| DNN-based | FlowNetS [6] | 21.56 | 7.30 | 11.44 | 43.65 | 135.79 | 21.60 | 7.31 | 11.19 | 44.38 | 139.85 |
| | FlowNet 2.0 [14] | 9.68 | 1.62 | 4.56 | 18.10 | 87.71 | 9.83 | 1.58 | 4.53 | 18.90 | 92.40 |
| | PWC-Net [27] | 10.07 | 1.75 | 4.67 | 19.72 | 89.12 | 10.65 | 1.71 | 4.71 | 21.14 | 100.46 |
| | IRR [13] | 9.51 | 1.84 | 4.63 | 19.05 | 79.18 | 9.95 | 1.81 | 4.70 | 19.75 | 88.94 |
| | MaskFlow(ours) | 9.26 | 1.35 | 3.51 | 17.2 | 89.82 | 11.20 | 1.82 | 5.02 | 22.51 | 103.74 |

Table 2: Comparison of AEE on the FlyingThings3D Clean dataset [20].

Runtime. The runtime of MaskFlow was on average 280ms per frame on a 512x512 input, using a computer with a single GTX 2080 Ti GPU. The majority of this time (200ms) was occupied by the Mask R-CNN-based network. We note that an extensive optimisation of the architecture could improve MaskFlow's efficiency significantly.

# 6 Conclusion

We introduced a novel approach for incorporating object-level representations for motion estimation. Our method integrates both object and dense correspondence matching in one framework. We perform on par with state of the art methods on conventional motion estimation datasets but significantly outperforms them in very challenging scenarios. Our experiments show that we can produce a more accurate motion field when fast-moving objects undergo a drastic change in their appearance. This widens the range of applications for our motion estimator over the previous state of the art, enabling it to address these challenges, which exist in low-framerate video and fast-moving scenes, more effectively. We see this approach as continuing the trend of motion estimators to exploit increasingly high-level scene understanding to provide more accurate and robust results. Future approaches might continue this trend, by estimating the 3D structure of objects to provide more accurate estimates of the rotational component of the motion field.

# References

[1] Aria Ahmadi and Ioannis Patras. Unsupervised convolutional neural networks for motion estimation. In *2016 IEEE international conference on image processing (ICIP)*, pages 1629–1633. IEEE, 2016.

[2] Aria Ahmadi, Ioannis Marras, and Ioannis Patras. Likenet: A siamese motion estimation network trained in an unsupervised way. In *BMVC*, page 296, 2018.

[3] Thomas Brox and Jitendra Malik. Large displacement optical flow: descriptor matching in variational motion estimation. *IEEE transactions on pattern analysis and machine intelligence*, 33(3):500–513, 2010.

[4] Thomas Brox, Andrés Bruhn, Nils Papenberg, and Joachim Weickert. High accuracy optical flow estimation based on a theory for warping. In *Computer Vision-ECCV 2004*, pages 25–36. Springer, 2004.

[5] Ricardo JGB Campello, Davoud Moulavi, Arthur Zimek, and Jörg Sander. Hierarchical density estimates for data clustering, visualization, and outlier detection. *ACM Transactions on Knowledge Discovery from Data (TKDD)*, 10(1):5, 2015.

[6] Alexey Dosovitskiy, Philipp Fischer, Eddy Ilg, Philip Hausser, Caner Hazirbas, Vladimir Golkov, Patrick Van Der Smagt, Daniel Cremers, and Thomas Brox. Flownet: Learning optical flow with convolutional networks. In *Proceedings of the IEEE international conference on computer vision*, pages 2758–2766, 2015.

[7] Blender Foundation. *Cycles Rendering Engine*. https://www.cycles-renderer.org/.

[8] Kaiming He, Georgia Gkioxari, Piotr Dollár, and Ross Girshick. Mask r-cnn. In *Proceedings of the IEEE international conference on computer vision*, pages 2961–2969, 2017.

[9] Berthold KP Horn and Brian G Schunck. Determining optical flow. *Artificial intelligence*, 17(1-3):185–203, 1981.

[10] Yinlin Hu, Rui Song, and Yunsong Li. Efficient coarse-to-fine patchmatch for large displacement optical flow. In *Proceedings of the IEEE Conference on Computer Vision and Pattern Recognition*, pages 5704–5712, 2016.

[11] Zhaojin Huang, Lichao Huang, Yongchao Gong, Chang Huang, and Xinggang Wang. Mask scoring r-cnn. In *The IEEE Conference on Computer Vision and Pattern Recognition (CVPR)*, June 2019.

[12] Tak-Wai Hui, Xiaoou Tang, and Chen Change Loy. Liteflownet: A lightweight convolutional neural network for optical flow estimation. In *Proceedings of the IEEE Conference on Computer Vision and Pattern Recognition*, pages 8981–8989, 2018.

[13] Junhwa Hur and Stefan Roth. Iterative residual refinement for joint optical flow and occlusion estimation. In *Proceedings of the IEEE Conference on Computer Vision and Pattern Recognition*, pages 5754–5763, 2019.

[14] Eddy Ilg, Nikolaus Mayer, Tonmoy Saikia, Margret Keuper, Alexey Dosovitskiy, and Thomas Brox. Flownet 2.0: Evolution of optical flow estimation with deep networks. In *Proceedings of the IEEE conference on computer vision and pattern recognition*, pages 2462–2470, 2017.

[15] J Yu Jason, Adam W Harley, and Konstantinos G Derpanis. Back to basics: Unsupervised learning of optical flow via brightness constancy and motion smoothness. In *European Conference on Computer Vision*, pages 3–10. Springer, 2016.

[16] Ivan Krasin, Tom Duerig, Neil Alldrin, Vittorio Ferrari, Sami Abu-El-Haija, Alina Kuznetsova, Hassan Rom, Jasper Uijlings, Stefan Popov, Shahab Kamali, Matteo Malloci, Jordi Pont-Tuset, Andreas Veit, Serge Belongie, Victor Gomes, Abhinav Gupta, Chen Sun, Gal Chechik, David Cai, Zheyun Feng, Dhyanesh Narayanan, and Kevin Murphy. Openimages: A public dataset for largescale multi-label and multi-class image classification. *Dataset available from https://storage.googleapis.com/openimages/web/index.html*, 2017.

[17] Pengpeng Liu, Michael R. Lyu, Irwin King, and Jia Xu. Selflow: Self-supervised learning of optical flow. In *CVPR*, 2019.

[18] Bruce D Lucas, Takeo Kanade, et al. An iterative image registration technique with an application to stereo vision. 1981.

[19] D. Maurer and A. Bruhn. Proflow: Learning to predict optical flow. In *BMVC*, 2018.

[20] N. Mayer, E. Ilg, P. Häusser, P. Fischer, D. Cremers, A. Dosovitskiy, and T. Brox. A large dataset to train convolutional networks for disparity, optical flow, and scene flow estimation. In *IEEE International Conference on Computer Vision and Pattern Recognition (CVPR)*, 2016. URL http://lmb.informatik.uni-freiburg.de/Publications/2016/MIFDB16. arXiv:1512.02134.

[21] Simon Meister, Junhwa Hur, and Stefan Roth. Unflow: Unsupervised learning of optical flow with a bidirectional census loss. In *Thirty-Second AAAI Conference on Artificial Intelligence*, 2018.

[22] Michal Neoral, Jan Šochman, and Ji ˇrí Matas. Continual occlusion and optical flow estimation. In *Asian Conference on Computer Vision*, pages 159–174. Springer, 2018.

[23] Anurag Ranjan and Michael J Black. Optical flow estimation using a spatial pyramid network. In *Proceedings of the IEEE Conference on Computer Vision and Pattern Recognition*, pages 4161–4170, 2017.

[24] Zhile Ren, Orazio Gallo, Deqing Sun, Ming-Hsuan Yang, Erik B Sudderth, and Jan Kautz. A fusion approach for multi-frame optical flow estimation. In *Proceedings of the IEEE Winter Conference on Applications of Computer Vision (WACV)*, 2019.

[25] Jerome Revaud, Philippe Weinzaepfel, Zaid Harchaoui, and Cordelia Schmid. Epicflow: Edge-preserving interpolation of correspondences for optical flow. In *Proceedings of the IEEE conference on computer vision and pattern recognition*, pages 1164–1172, 2015.

[26] Florian Schroff, Dmitry Kalenichenko, and James Philbin. Facenet: A unified embedding for face recognition and clustering. In *Proceedings of the IEEE conference on computer vision and pattern recognition*, pages 815–823, 2015.

[27] Deqing Sun, Xiaodong Yang, Ming-Yu Liu, and Jan Kautz. Pwc-net: Cnns for optical flow using pyramid, warping, and cost volume. In *Proceedings of the IEEE Conference on Computer Vision and Pattern Recognition*, pages 8934–8943, 2018.

[28] Deqing Sun, Xiaodong Yang, Ming-Yu Liu, and Jan Kautz. Models matter, so does training: An empirical study of cnns for optical flow estimation. *IEEE Transactions on Pattern Analysis and Machine Intelligence*, 2019.

[29] Robert L Thorndike. Who belongs in the family. In *Psychometrika*. Citeseer, 1953.

[30] Yang Wang, Yi Yang, Zhenheng Yang, Liang Zhao, Peng Wang, and Wei Xu. Occlusion aware unsupervised learning of optical flow. In *Proceedings of the IEEE Conference on Computer Vision and Pattern Recognition*, pages 4884–4893, 2018.

[31] Philippe Weinzaepfel, Jerome Revaud, Zaid Harchaoui, and Cordelia Schmid. Deepflow: Large displacement optical flow with deep matching. In *Proceedings of the IEEE international conference on computer vision*, pages 1385–1392, 2013.

[32] Philippe Weinzaepfel, Jerome Revaud, Zaid Harchaoui, and Cordelia Schmid. DeepFlow: Large displacement optical flow with deep matching. In *IEEE Intenational Conference on Computer Vision (ICCV)*, Sydney, Australia, December 2013. URL http://hal.inria.fr/hal-00873592.

[33] Jonas Wulff and Michael J Black. Efficient sparse-to-dense optical flow estimation using a learned basis and layers. In *Proceedings of the IEEE Conference on Computer Vision and Pattern Recognition*, pages 120–130, 2015.

[34] Jonas Wulff, Laura Sevilla-Lara, and Michael J Black. Optical flow in mostly rigid scenes. In *Proceedings of the IEEE Conference on Computer Vision and Pattern Recognition*, pages 4671–4680, 2017.

# Supplementary Material for "MaskFlow: Object-Aware Motion Estimation"

Aria Ahmadi
aria.ahmadi@imgtec.com
David R. Walton
david.walton@imgtec.com
Tim Atherton
tim.atherton@imgtec.com
Cagatay Dikici
cagatay.dikici@imgtec.com

Imagination Technologies
Kings Langley
Hertfordshire, UK

## 7 Overview

In this supplementary material, we present additional results to complement the main manuscript. First, we provided an example of Maskflow Dataset generation process in Section 2. Second, we detailed object instance matching parameters as well as visual examples of object instance matching in Section 3. Third, we illustrate a translation motion field example in Section 4.

## 8 Rendering Maskflow Dataset

Figure 1 shows an example of rendering a single sample for Maskflow dataset. The 3D scene is shown, as well as the ground truth data we render. This consists of the two frames themselves, a translation motion field, a full motion field and an index map.

In the top row of the figure, we show 3D preview renders of the scene for the reference and target frames, viewed from a different angle. Note that the objects and background plane both move. The remaining rows show the outputs of our rendering process. The upper middle row contains the colour reference and target frames. The lower middle row shows the translation flow field (left) and the full flow field (right). The final row shows index maps for the reference and target frames.

Color Coding. To visualize the estimated motion field, the direction and magnitude of the displacements are encoded according to a color wheel and by the saturation respectively, as illustrated in figure. 2.

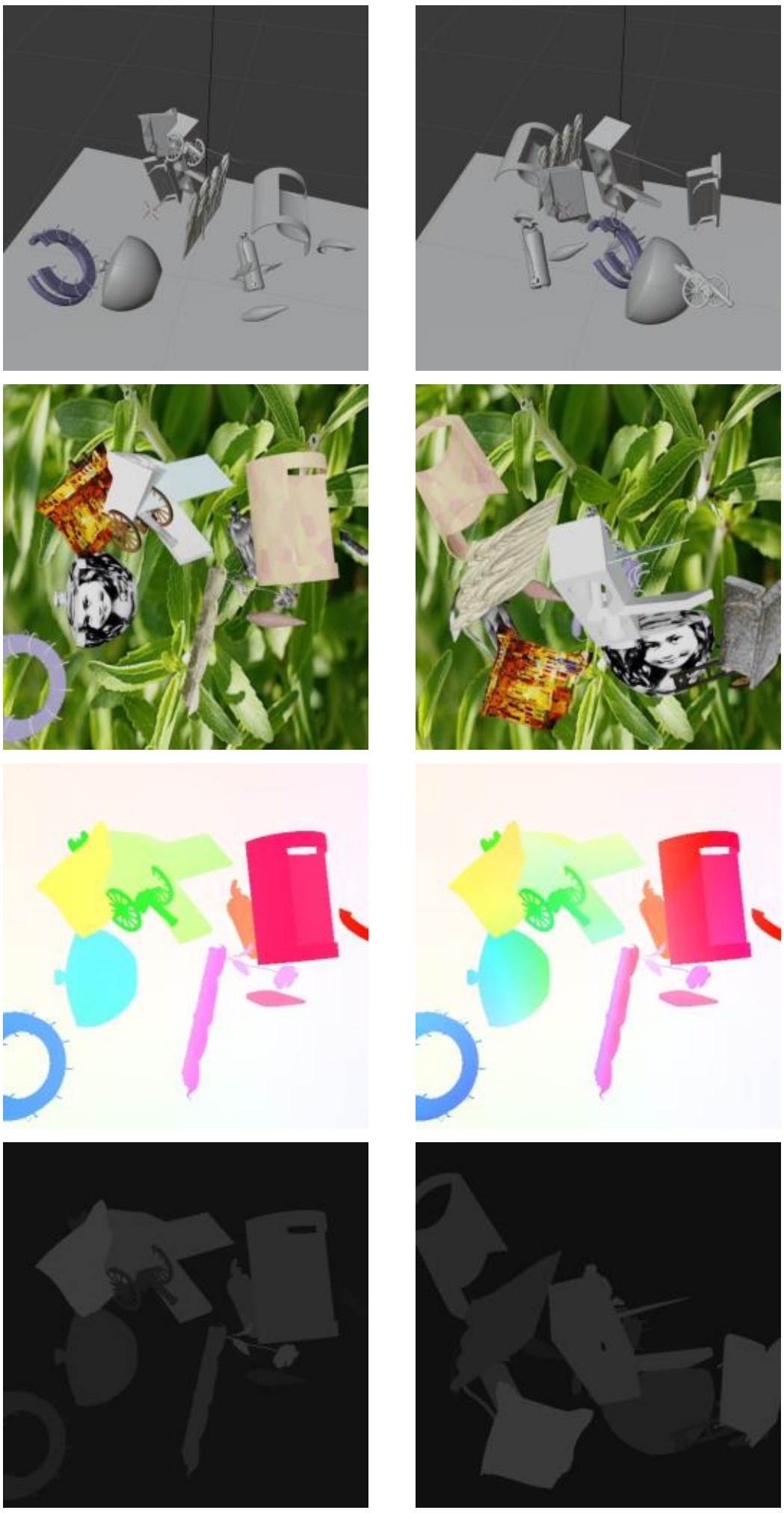

Figure 1: Example of rendering a sample from our dataset.

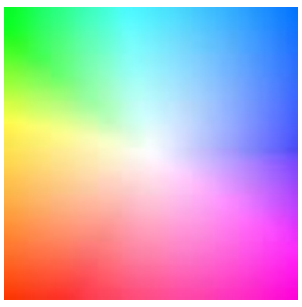

Figure 2: The color wheel used for visualization of the motion fields in this paper. The displacement of every pixel in this figure is the vector from the center of the square to this pixel. The central pixel does not move. This visualisation method is similar to that used in [5]. But unlike in [5], the values are scaled based on the maximum motion magnitude in the ground truth to provide a visual comparison of different methods' ability of estimation in a challenging setting.

# 9 Object Instance Matching Parameters and Examples

Figure 3 shows some examples of the output of our first segmentation and matching step. We show the final matched candidates only. The masks are highlighted and matched candidates are connected with red lines. As illustrated in figure 3 (a), the matching step has a good performance. Although a small proportion of the objects are not identified, the refiner may be able to obtain motion vectors for them. Figure 3 (b) shows an example of imperfect segmentation of an object. Figure 3 (c) shows an example of appearance of similar objects in one scene. This is a very difficult scenario where it is difficult even for humans to identify the matching objects. Figure 3 (d) shows an example of wrongly matched objects (mismatches exist but are not very often).

We note that in (a) the matcher is able to match the small, fast-moving object highlighted in green at the bottom of the images, although this object match would typically be missed by a pyramid-based motion estimator. An example for this is illustrated in figure 5. At the top left of the reference frame, there is a small, fast-moving object highlighted with a red rectangle. We note that only MaskFlow was able to correctly distinguish this object from the one behind it and generate correct motion vectors for it in this case.

Matching Parameters. The parameters we used for matching the objects are provided in table 1:

| Dataset | Min Area | Mask Score | Objectness Score | Cluster Compactness | Min Distance $d$ |
|---|---|---|---|---|---|
| MaskFlow | 1500 | 0.9 | 0.9 | 4000 | 2 |
| FlyingThings3D | 2500 | 0.95 | 0.95 | 2000 | 2.5 |

Table 1: The parameters we obtained from our grid search for the matching part of the method.

In table 1, cluster compactness is the sum of squared distances of data items from the nearest cluster center after Kmeans terminates (this is sometimes referred to as Kmeans inertia). Min area is the threshold applied to the area of each candidate's mask, in pixels. Min distance is the final threshold applied to the feature distance between potential matching candidate pairs.

# 10 Translation Motion Field Example

Figure 4 shows an example visualisation of a full flow field $D$ and the object translation-only flow field $D_t$, as defined in section 3 of the main paper. Note that the translation flow field $D_t$ is constant within the boundary of each object, whereas the full flow field $D$ varies within each object due to rotation and perspective projection effects.

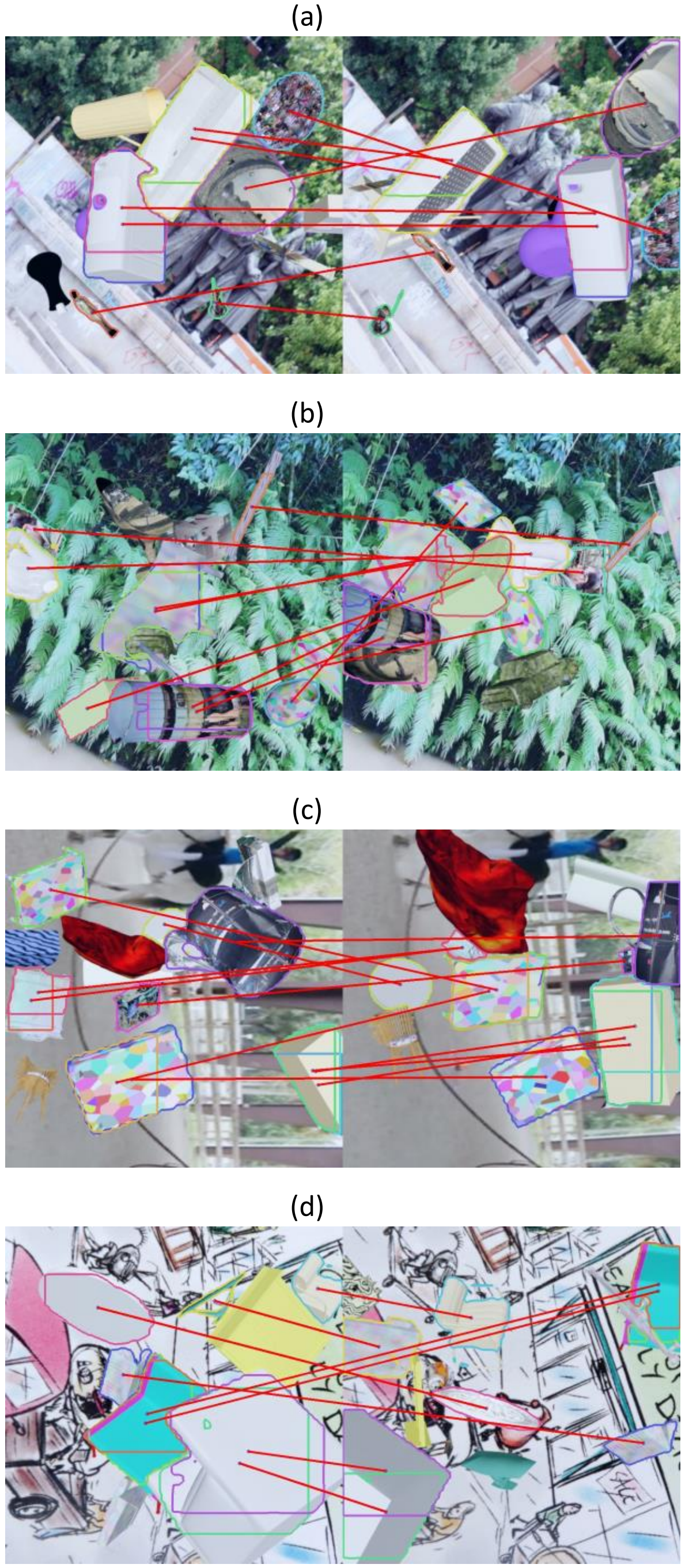

Figure 3: Example of the output of our matching step.

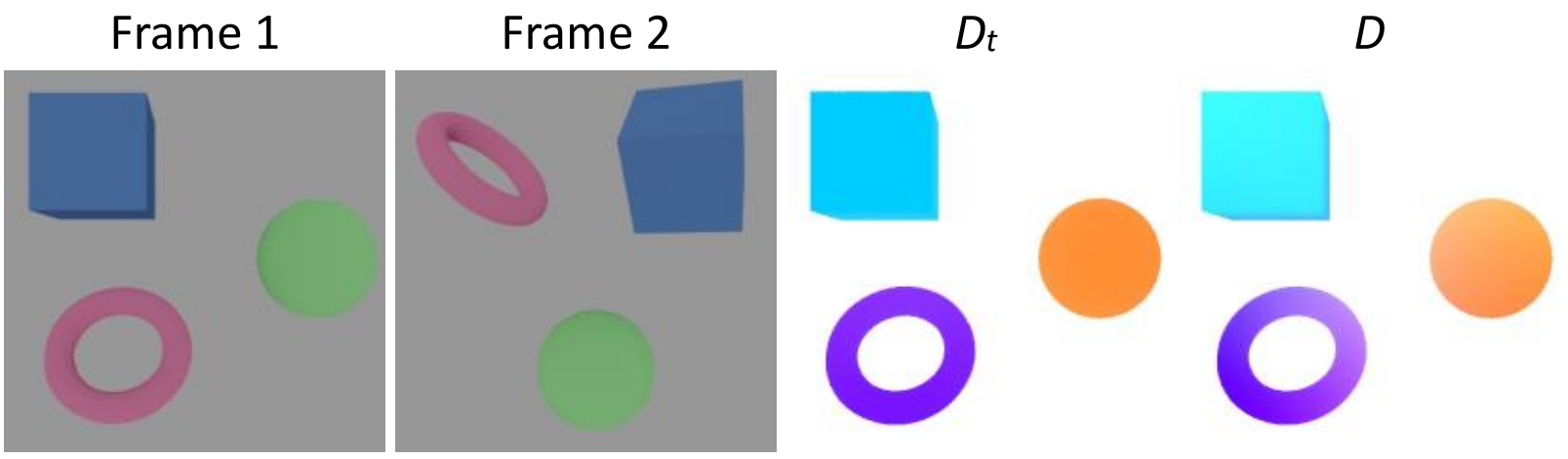

Figure 4: Example visualisations of the ground truth for the translation-only motion field $D_t$ and full motion field $D$.

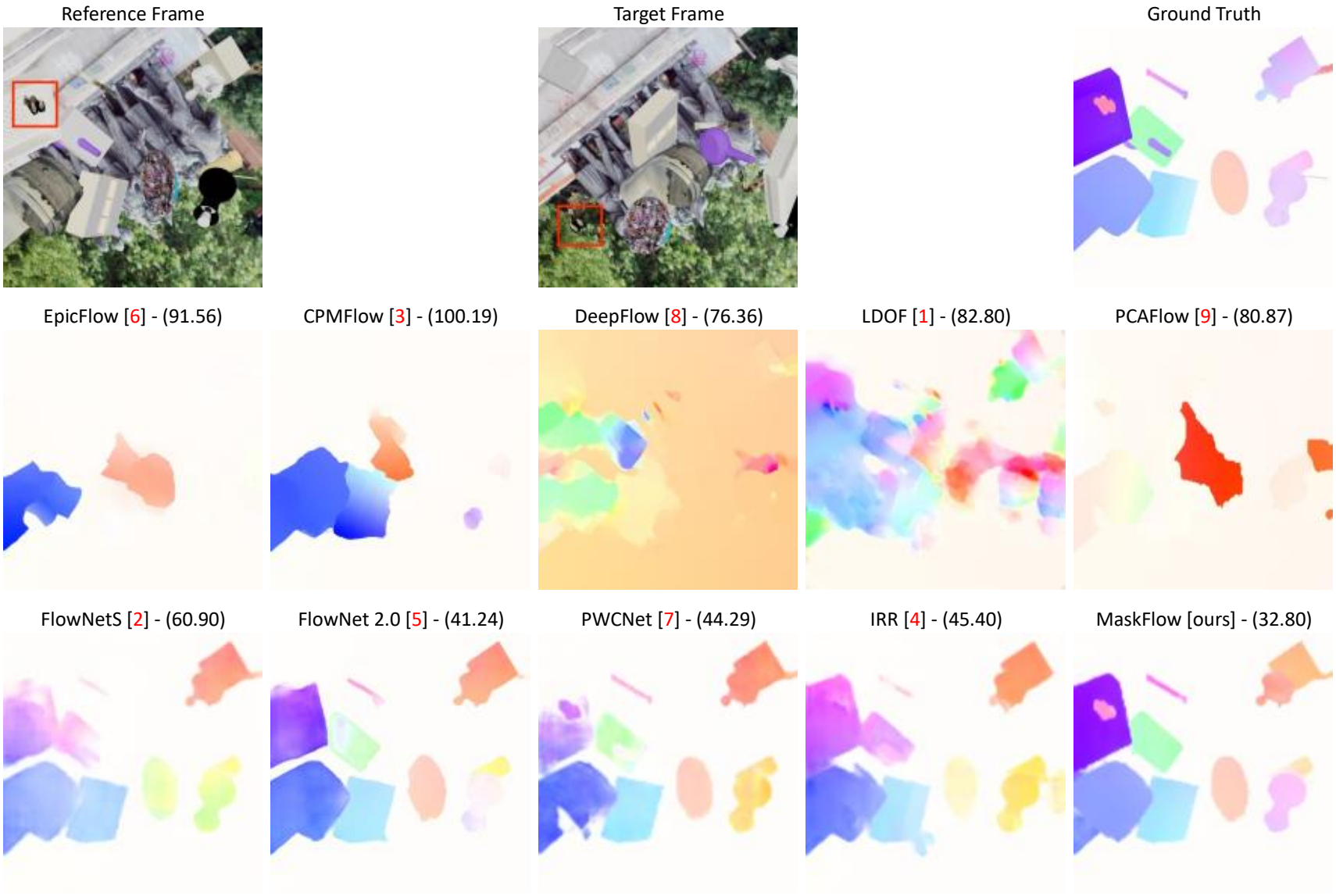

Figure 5: The input, ground truth, and output of SOA classic and DNN-based methods on one sample from the test split of our MaskFlow dataset - the parentheses contain the AEE in pixels. For a fair comparison, we have trained all DNN-based methods under the same training schemes, $S_{long}$ and $S_{fine}$, proposed in [5].

# References

[1] Thomas Brox and Jitendra Malik. Large displacement optical flow: descriptor matching in variational motion estimation. *IEEE transactions on pattern analysis and machine intelligence*, 33(3):500–513, 2010.

[2] Alexey Dosovitskiy, Philipp Fischer, Eddy Ilg, Philip Hausser, Caner Hazirbas, Vladimir Golkov, Patrick Van Der Smagt, Daniel Cremers, and Thomas Brox. Flownet: Learning optical flow with convolutional networks. In *Proceedings of the IEEE international conference on computer vision*, pages 2758–2766, 2015.

[3] Yinlin Hu, Rui Song, and Yunsong Li. Efficient coarse-to-fine patchmatch for large displacement optical flow. In *Proceedings of the IEEE Conference on Computer Vision and Pattern Recognition*, pages 5704–5712, 2016.

[4] Junhwa Hur and Stefan Roth. Iterative residual refinement for joint optical flow and occlusion estimation. In *Proceedings of the IEEE Conference on Computer Vision and Pattern Recognition*, pages 5754–5763, 2019.

[5] Eddy Ilg, Nikolaus Mayer, Tonmoy Saikia, Margret Keuper, Alexey Dosovitskiy, and Thomas Brox. Flownet 2.0: Evolution of optical flow estimation with deep networks. In *Proceedings of the IEEE conference on computer vision and pattern recognition*, pages 2462–2470, 2017.

[6] Jerome Revaud, Philippe Weinzaepfel, Zaid Harchaoui, and Cordelia Schmid. Epicflow: Edge-preserving interpolation of correspondences for optical flow. In *Proceedings of the IEEE conference on computer vision and pattern recognition*, pages 1164–1172, 2015.

[7] Deqing Sun, Xiaodong Yang, Ming-Yu Liu, and Jan Kautz. Pwc-net: Cnns for optical flow using pyramid, warping, and cost volume. In *Proceedings of the IEEE Conference on Computer Vision and Pattern Recognition*, pages 8934–8943, 2018.

[8] Philippe Weinzaepfel, Jerome Revaud, Zaid Harchaoui, and Cordelia Schmid. Deepflow: Large displacement optical flow with deep matching. In *Proceedings of the IEEE international conference on computer vision*, pages 1385–1392, 2013.

[9] Jonas Wulff and Michael J Black. Efficient sparse-to-dense optical flow estimation using a learned basis and layers. In *Proceedings of the IEEE Conference on Computer Vision and Pattern Recognition*, pages 120–130, 2015.